\documentclass[mlmain]{jmlr}

\usepackage{longtable}% for long tables

\usepackage{booktabs}
\usepackage[load-configurations=version-1]{siunitx} % newer version
\theorembodyfont{\upshape}
\theoremheaderfont{\scshape}
\theorempostheader{:}
\theoremsep{\newline}

\DeclareMathOperator{\im}{im}

\jmlrvolume{}
\firstpageno{1}
\jmlryear{2025}
\jmlrworkshop{Symmetry and Geometry in Neural Representations}

\title[Sheaf Cohomology of Linear PC]{Sheaf Cohomology of Linear Predictive Coding Networks}

\author{\Name{Jeffrey Seely} \Email{jeffrey@sakana.ai}\\
\addr Sakana AI
}

\begin{document}

\maketitle

\begin{abstract}
Predictive coding (PC) replaces global backpropagation with local optimization over weights and activations. We show that linear PC networks admit a natural formulation as cellular sheaves: the sheaf coboundary maps activations to edge-wise prediction errors, and PC inference is diffusion under the sheaf Laplacian. Sheaf cohomology then characterizes irreducible error patterns that inference cannot remove. We analyze recurrent topologies where feedback loops create internal contradictions, introducing prediction errors unrelated to supervision. Using a Hodge decomposition, we determine when these contradictions cause learning to stall. The sheaf formalism provides both diagnostic tools for identifying problematic network configurations and design principles for effective weight initialization for recurrent PC networks.
\end{abstract}


\section{Introduction}
\label{sec:intro}

Predictive coding (PC) recasts neural network training as an optimization problem over weights and \emph{activations} \citep{Salvatori2025}. This replaces a global loss with a series of layer-local losses. Optimizing over activations (PC inference) enables one to handle arbitrary recurrent topologies without unrolling the computational graph or relying on backpropagation through time \citep{Salvatori2022}. But it creates a fundamental ambiguity that, to our knowledge, has been overlooked: a node deep in a recurrent network receives error signals from all its connections---some from supervision, others from contradictory feedback loops---with no way to distinguish between them.

We formalize linear PC networks as cellular sheaves \citep{Curry2014-a} to analyze this problem systematically. A \emph{predictive coding sheaf} assigns vector spaces to vertices (neuron activations) and edges (prediction errors) of the underlying computational graph, with restriction maps encoding the network weights. The sheaf coboundary operator $\delta^0$ computes all prediction errors simultaneously. Sheaf cohomology represents the network's capacity to resolve inconsistencies. At a high level, sheaf theory studies how local consistency patches together into global coherence \citep{Rosiak2022-a}---detecting when this local-to-global transfer \emph{succeeds} (captured by the zeroth cohomology $H^0$) and when it \emph{fails} (captured by $H^1$). This theme matches the predictive coding ambition: how can locally informed layers coordinate to solve a global task, even amid feedback loops far from any supervision?

We illustrate the correspondence between predictive coding networks and sheaf theory through simple examples (\sectionref{sec:two-layer,sec:sheaf}). When input and output nodes are clamped to data and supervision targets, a Hodge decomposition characterizes precisely how the supervision signal distributes prediction errors across edges, and how it distributes neuron activations across vertices (\sectionref{sec:relative}). This framework reveals why certain recurrent networks fail to learn: feedback loops can create internal contradictions that dominate supervision signals. Experiments demonstrate orders-of-magnitude differences in learning convergence rates based on weight initialization (\sectionref{sec:examples}); importantly, the stark difference is not due to weight magnitude (all our initializations are orthonormal) but due to how the orientations of the weights create \emph{tension} or \emph{resonance} when traced around feedback loops within the network. We focus on linear networks (where sheaf cohomology and Hodge decompositions can be readily computed), though what follows can be regarded as a linear analysis of nonlinear PC networks evaluated at equilibria.

\begin{figure}[tbp]
   \floatconts
    {fig:sheaf}
    {\caption{Cellular sheaf over a graph. Vector spaces are assigned to nodes and edges, linear restriction maps are assigned to vertex-to-edge pairs. Layers $h_i=W_i h_{i-1}$ are encoded via a restriction pair $(W_i, I)$. This sheaf represents the deep linear network $y=W_3 W_2 W_1 x$.}}
    {\includegraphics[width=0.85\linewidth]{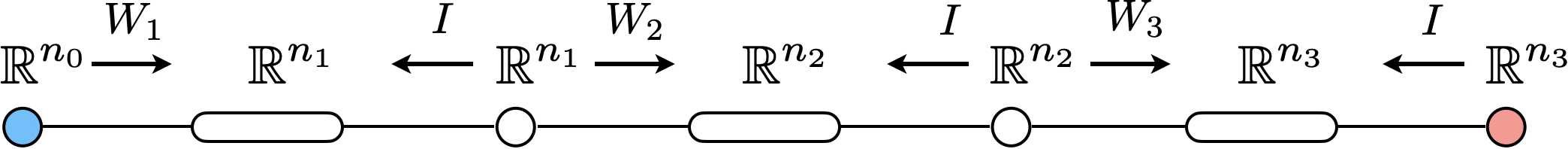}}
\end{figure}

\section{Deep Linear Networks as Constraints and Penalties}
\label{sec:two-layer}

We will establish the connection between linear PC networks and cellular sheaves by way of example.

\subsection{A three-layer warm-up}
Let $x \in \mathbb{R}^{n_0}$, $h_1 \in \mathbb{R}^{n_1}$, $h_2 \in \mathbb{R}^{n_2}$, and $y \in \mathbb{R}^{n_3}$. A three-layer deep linear network optimizes
\begin{equation}
    \label{eq:unconstrained}
    \min_{W_1, W_2, W_3}\; \tfrac12\,\big\lVert y - W_3 W_2 W_1 x\big\rVert_2^2\,.
\end{equation}
Alternatively, we can write \equationref{eq:unconstrained} as a constrained optimization problem,
\begin{align}
    \label{eq:constrained}
    \min_{W_1, W_2, W_3, h_1, h_2}\; \tfrac12\,\big\lVert y - W_3 h_2 \big\rVert_2^2 \quad
    \text{s.t.}\quad h_1=W_1x,\quad h_2=W_2 h_1,
\end{align}
whereby the key feature is that now $h_1$ and $h_2$ are optimization variables. When constraints are satisfied exactly, the problem is equivalent to \equationref{eq:unconstrained}.

\subsection{Predictive coding as a penalty method}
One interpretation of predictive coding is that it is a soft relaxation of \equationref{eq:constrained} via quadratic regularization terms, otherwise known as a penalty method.\footnote{To our knowledge, PC as a penalty method is not standard pedagogy, though we find this perspective instructive for what follows.} The resulting loss function, called the total energy $E_{\text{PC}}$, is the sum of squared residuals (or prediction errors):
\begin{align}
    \label{eq:energy}
    E_{\text{PC}}(h_1, h_2, W_1, W_2, W_3) = \tfrac12\big(\lVert r_1\rVert_2^2+\lVert r_2\rVert_2^2+\lVert r_3\rVert_2^2\big),\\
    r_1 = h_1 - W_1 x,\quad
    r_2 = h_2 - W_2 h_1,\quad
    r_3 = y - W_3 h_2.
\end{align}
PC is a bilevel optimization problem: \textbf{Inference} fixes $W_i$s and minimizes $E_{\text{PC}}$ over $h_i$s; \textbf{learning} then fixes $h_i$s and updates $W_i$s. 
In practice, inference amounts to multiple gradient descent steps (w.r.t. $h_i$s) and learning is a single gradient descent step (w.r.t. $W_i$s) before $h_i$s are initialized again for the next minibatch; see \citet{Salvatori2022-a} for other update schedules.

\section{Sheaf Preliminaries for Predictive Coding}
\label{sec:sheaf}
We formalize a linear PC network as a cellular sheaf \citep{Curry2014-a}, establishing nomenclature along the way.

\paragraph{From activations to cochains.}
Following the three-layer example, we organize (concatenate) activations into a \textbf{0-cochain}:
\begin{align}
    s = (x, h_1, h_2, y) \in C^0 := \mathbb{R}^{n_0} \oplus \mathbb{R}^{n_1} \oplus \mathbb{R}^{n_2} \oplus \mathbb{R}^{n_3}.
\end{align}
The vector-concatenation of prediction errors forms a \textbf{1-cochain}:
\begin{align}
    r = (r_1, r_2, r_3) \in C^1 := \mathbb{R}^{n_1} \oplus \mathbb{R}^{n_2} \oplus \mathbb{R}^{n_3}.
\end{align}

\paragraph{The coboundary operator.}
The coboundary $\delta^0: C^0 \to C^1$ computes all prediction errors simultaneously from the activations:
\begin{align}
    \label{eq:coboundary}
    \delta^0 s = \begin{bmatrix}
    -W_1 & I_{n_1} & 0 & 0 \\
    0 & -W_2 & I_{n_2} & 0 \\
    0 & 0 & -W_3 & I_{n_3}
    \end{bmatrix}
    \begin{bmatrix} x \\ h_1 \\ h_2 \\ y \end{bmatrix}
    = \begin{bmatrix} h_1 - W_1 x \\ h_2 - W_2 h_1 \\ y - W_3 h_2 \end{bmatrix},
\end{align}
where we note the per-edge sign convention.

\paragraph{PC Inference is sheaf diffusion.}
From \equationref{eq:coboundary} it is clear that PC energy \equationref{eq:energy} can be written compactly as $E_{\text{PC}}(s) = \tfrac{1}{2}\|\delta^0 s\|^2$. Minimizing the energy $E_{\text{PC}}$ with respect to $s$ via gradient descent yields the gradient flow $\dot{s} = -L s$ (or its discrete-time variant), where $L=(\delta^0)^\top \delta^0$ is the \textbf{sheaf Laplacian}. The gradient flow is known as \textbf{sheaf diffusion}, as it generalizes notions of graph diffusion to sheaves \citep{Hansen2018-aq}.

\paragraph{Network sheaves.}
More generally, for an arbitrary graph $G = (V, E)$, a \emph{network sheaf} $\mathcal{F}$ over $G$ assigns vector spaces $\mathcal{F}(v) = \mathbb{R}^{n_v}$ to vertices and $\mathcal{F}(e) = \mathbb{R}^{m_e}$ to edges. Each edge $e = (u, v)$ carries linear restriction maps $\rho_{e \leftarrow u}: \mathcal{F}(u) \to \mathcal{F}(e)$ and $\rho_{e \leftarrow v}: \mathcal{F}(v) \to \mathcal{F}(e)$.

Generalizing the three-layer example from the previous section, we define a \emph{predictive coding sheaf} to be a network sheaf with the following convention: $e=(u \to v)$ denotes a forward computation $s_v = W_e s_u$, encoded via restrictions $\rho_{e \leftarrow u} = W_e$ and $\rho_{e \leftarrow v} = I$. The coboundary then computes the prediction error at each edge:
\begin{align}
(\delta^0 s)_e = \rho_{e \leftarrow v} s_v - \rho_{e \leftarrow u} s_u = s_v - W_e s_u.
\end{align}
This definition applies for a graph $G$ with multiple edges between vertices. Formally, our graph is undirected but where we use notation $e=(u\to v)$ to specify the placement of the $W_e$ and $I$ restriction maps.

\paragraph{Cohomology.}
The sheaf coboundary induces two cohomology groups that characterize network consistency:
\begin{align}
H^0(G, \mathcal{F}) &= \ker \delta^0 = \{s \in C^0 : \delta^0 s = 0\}, \\
H^1(G, \mathcal{F}) &= C^1 / \im \delta^0 \cong \ker (\delta^0)^\top.
\end{align}
Here $H^0$ consists of activation patterns (on vertices) with zero prediction error everywhere, while $H^1$ represents prediction error patterns (on edges) that cannot arise from any activation choice. In finite dimensions with inner products we can represent $H^1(G,\mathcal{F})$ by
$\ker (\delta^0)^\top$. In our setting the predictive coding sheaf $\mathcal{F}$
depends on the current network weights, so $H^0$ and $H^1$ should be understood
as weight-dependent linear subspaces of $C^0$ and $C^1$, not as topological
invariants of the underlying graph.

% In finite dimensions with inner products, $H^1 \cong \ker (\delta^0)^\top$.%We use elements of $\ker (\delta^0)^\top$ to represent elements of $H^1$.  In our setting the predictive coding sheaf $\mathcal{F}$ itself depends on the current network weights, so $H^0$ and $H^1$ should be understood as weight-dependent linear subspaces rather than topological invariants of the bare graph.

\section{Relative Systems and Clamped Networks}
\label{sec:relative}
In predictive coding, one typically does not work with the full unconstrained system. Instead, one clamps input $x$ to data and output $y$ to supervision targets during network training. In sheaf terms, this corresponds to a relative system. Let $V = V_{\text{free}} \cup V_{\text{clamped}}$ partition the vertices. The relative coboundary $D$ extracts the columns of $\delta^0$ corresponding to free vertices. In the three-layer network, we have:
\begin{align}
    D = \begin{bmatrix}
    I_{n_1} & 0 \\
    -W_2 & I_{n_2} \\
    0 & -W_3
    \end{bmatrix}, \quad
    b = \begin{bmatrix}
    -W_1 x \\
    0 \\
    y
    \end{bmatrix}.
\end{align}
During inference, we optimize only over internal activations $z = (h_1, h_2) \in \mathbb{R}^{n_1} \oplus \mathbb{R}^{n_2}$. The effect of clamping data nodes can be interpreted as creating a ``target prediction error'' $b$ as a 1-cochain.

More precisely, clamping induces the inhomogeneous least-squares system
\begin{align}
\label{eq:rel_energy}
E_{\mathrm{rel}}(z)=\tfrac12\|Dz+b\|^2,
\quad
z\in C^0_{\mathrm{free}}.
\end{align}
When $D$ has full column rank (as in all our examples), the minimizer is unique: $z^\star = -D^\dagger b$, where ${\bullet}^\dagger$ indicates the pseudoinverse. More generally, the solution set is the affine space $\{z^\star + z_{\mathrm{null}} : z_{\mathrm{null}} \in \ker D\}$, and gradient descent from $z=0$ converges to the minimum-norm representative $z^\star$. At any minimizer, the residual satisfies
\begin{align}
    r^\star := Dz^\star+b \in \ker D^\top,
    \qquad
    D^\top r^\star=0.
\end{align}
Equivalently, the target $b$ admits an orthogonal (Hodge) decomposition
\begin{align}
    b = \underbrace{(-Dz^\star)}_{\in\,\im D} \;+\; \underbrace{r^\star}_{\in\,\ker D^\top}.
\end{align}
This decomposition makes explicit the part of the ``target prediction error'' $b$ that can be eliminated by inference ($-Dz^\star$) and the part that cannot ($r^\star$).

\paragraph{Harmonic and diffusive operators.}
The orthogonal decomposition of $b$ allows us to precisely state how $b$ influences edge patterns and activation patterns at optimal inference. Define the harmonic projector $\mathcal{H}$ and diffusive operator $\mathcal{G}$ as
\begin{align}
    \mathcal{H} := I - D D^\dagger,
    \qquad
    \mathcal{G} := L_{\mathrm{rel}}^\dagger D^\top = D^\dagger,
\end{align}
where $L_{\mathrm{rel}} = D^\top D$ is the relative sheaf Laplacian. Then the minimum-norm solution of \equationref{eq:rel_energy} yields the compact form,
\begin{align}
    \label{eq:hodge-rel}
    r^\star = \mathcal{H} b, 
    \qquad
    z^\star = - \mathcal{G} b,
    \qquad
    E_{\mathrm{rel}}(z^\star)=\tfrac12\big\lVert \mathcal{H}b\big\rVert_2^2.
\end{align}
We can also think of $b$ as an ``excitation vector'' for the network, or as boundary conditions. The projector $\mathcal{H}$ distributes that excitation across edges into the harmonic subspace of $C^1$, $\ker D^\top$. The operator $\mathcal{G}$ diffuses the same excitation to internal vertices, determining which sources are actually active after inference/diffusion has converged.

\paragraph{Learning requires harmonic-diffusive overlap.}
For a trainable edge $e=(u\!\to\!v)$ with weight $W_e$ and source activation $s_u$, the per-edge gradient is
\begin{align}
    \frac{\partial E_{\mathrm{rel}}}{\partial W_e}
    &=(W_e s_u - s_v)s_u^\top = -\, r_e\, s_u^\top.
\end{align}

Learning requires both a residual and a source. If either $r_e$ or $s_u$ vanishes, the update at edge $e$ is zero. At optimal inference, we can specify where in the network this occurs. Define the full activation vector $s^\star \in C^0$ with free vertices from $z^\star$ and clamped vertices at their fixed values: $(s^\star)_{\text{free}} = z^\star$, $(s^\star)_{\text{clamped}} = (x, y)$. Using \equationref{eq:hodge-rel},
\begin{align}
    \label{eq:edge-grad-rel}
    \frac{\partial E_{\mathrm{rel}}}{\partial W_e}\Big|_{z^\star}
    = -\, (\mathcal{H} b)_e\,(s^\star)_u^\top.
\end{align}
When $u$ is a free vertex, $(s^\star)_u = -(\mathcal{G} b)_u$, yielding
\begin{align}
    \frac{\partial E_{\mathrm{rel}}}{\partial W_e}\Big|_{z^\star}
    = \underbrace{(\mathcal{H} b)_e}_{\text{harmonic (edge)}}\;\underbrace{(\mathcal{G} b)_u^\top}_{\text{diffusive (vertex)}}.
\end{align}

\subsection{Relative cohomology}
To connect our clamped setting to sheaf theory more formally, we point the reader to \citet{Hansen2020-ff,Hansen2018-aq}. In brief: the clamped setting yields a relative cochain complex $(C^\bullet(G,V_{\text{clamped}};\mathcal{F}), D)$ with target $b$ defining a cohomology class $[b] \in H^1_{\text{rel}} \cong \ker D^\top$ with $r^\star = \mathcal{H}b$ its harmonic representation. In this language, PC inference solves a discrete Dirichlet problem---harmonically extending boundary values to the interior with $H^1_{\text{rel}}$ measuring obstructions to this extension.

\subsection{A minimal recurrent example}
\label{sec:sheaf-recurrent}
Consider augmenting our three-layer network with a feedback edge $h_2\!\to\!h_1$, creating a cycle. The coboundary and relative coboundary gain an additional row. The relative system is
\begin{align}
    D = \begin{bmatrix}
    I_{n_1} & 0 \\
    -W_2 & I_{n_2} \\
    0 & -W_3 \\
    I_{n_1} & -W_2^{\text{FB}}
    \end{bmatrix}, \quad
    b = \begin{bmatrix}
    -W_1 x \\
    0 \\
    y \\
    0
    \end{bmatrix}.
\end{align}
The cycle induces a monodromy $\Phi = W_2^{\text{FB}} W_2 : \mathbb{R}^{n_1} \to \mathbb{R}^{n_1}$. During inference, $h_1$ receives prediction errors from three sources: the forward path ($h_1 - W_1x$), the next layer ($h_2 - W_2h_1$), and the feedback loop ($h_1 - W_2^{\text{FB}} h_2$). Consider two extreme cases of the weight initialization of the system:

\paragraph{Resonance ($\Phi \approx I$).} 
In this case, the feedback loop reinforces itself---changes in $h_1$ return unchanged. We show below that this may introduce slow modes during inference. But once converged, the network has no internal contradictions to resolve.

\paragraph{Internal tension ($\Phi \approx -I$).} 
The feedback loop contradicts itself---changes in $h_1$ return negated. During inference, $h_1$ compromises between forward predictions and contradictory feedback. Learning thus requires resolving two tasks: match the supervision signal and resolve the internal contradiction.

How \emph{tension} and \emph{resonance} enter the harmonic and diffusion operators follows from their effect on the linear dependence of the block rows of $D$. While the algebraic story is tractable, we pursue empirical demonstration in the next section, using a 10-node network to concretely show how monodromy shapes these operators and ultimately learning itself.

\begin{figure}[!htbp]
    \floatconts
      {fig:tension}
      {\caption{Harmonic-diffusive separation in a 10-node network for three initializations.
      Top: sample-averaged harmonic load per edge.
      Middle: sample-averaged diffusive activation per vertex.
      Bottom: weight update magnitude. See \sectionref{sec:examples} for details.
      For $\theta=0$ (resonant), harmonic load and diffusion overlap broadly.
      For $\theta=\pi$ (contradictory), harmonic load concentrates on internal edges where diffusion does not reach, starving the corresponding updates.}
      }
      {\includegraphics[width=0.98\linewidth]{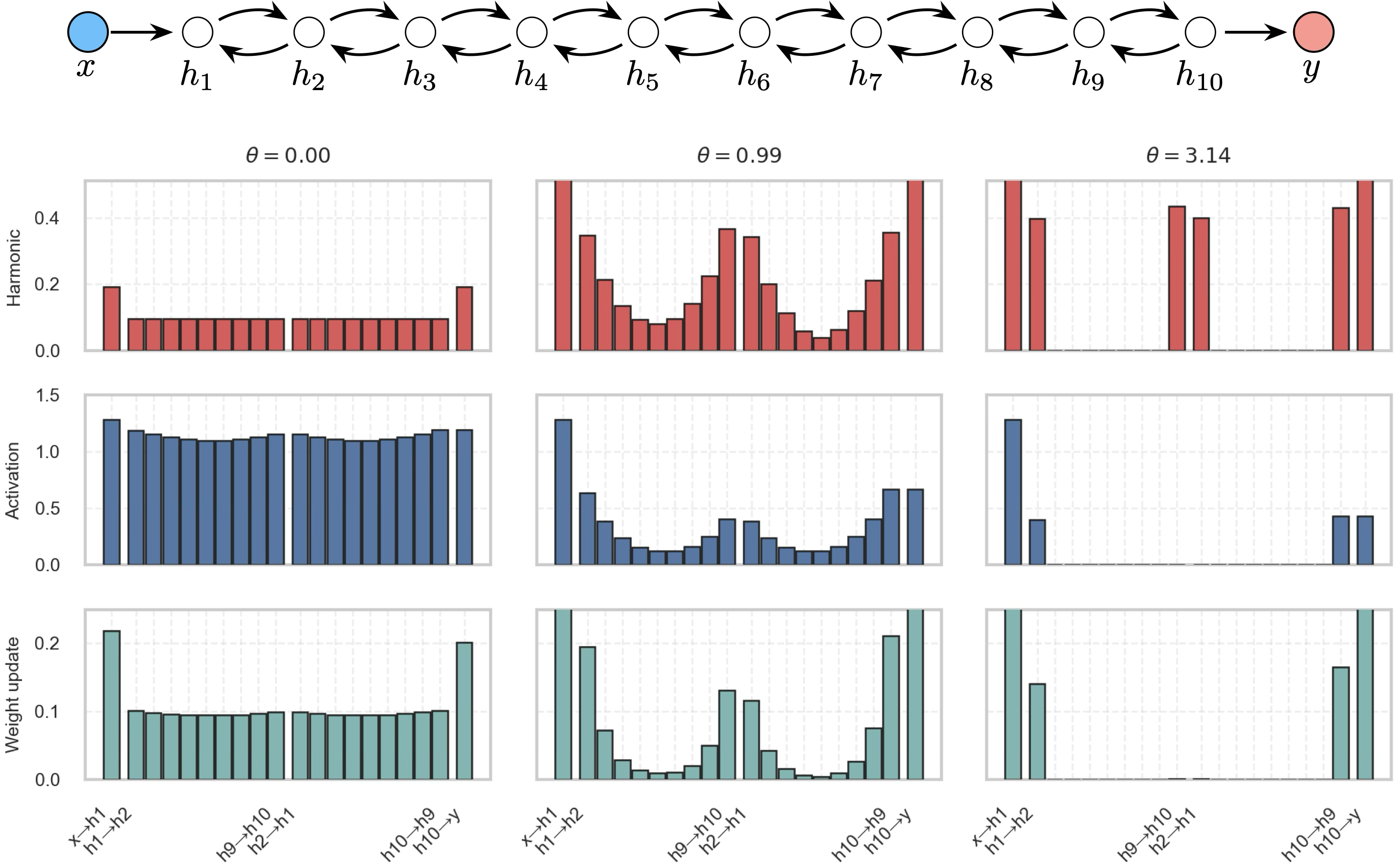}}
 \end{figure}

\section{A Canonical Knotted Network}
\label{sec:examples}

Consider the network in \figureref{fig:tension}: a 10-layer deep linear network with feedback cycles between each layer. We initialize forward weights $W_i$ as random $2\times 2$ orthonormal matrices. Feedback weights $W_i^{\text{FB}}$ are initialized so the monodromy $\Phi_i(\theta)=W_i^{\text{FB}}W_i$ induces a net rotation of $\theta$. Thus $\theta=0$ sets $W_i^{\text{FB}}=W_i^{-1}$ (resonance) and $\theta=\pi$ yields $\Phi_i(\theta)=-I$ (strong internal tension).

We run PC to disentangle the network and recover the identity mapping.
We draw samples $x\in\mathbb{R}^2$ from a standard normal and set $y=x$ (optionally with small i.i.d.\ Gaussian noise). Let $B = [b^{(1)}, \ldots, b^{(N)}]$ denote a batch of $N=128$ target vectors. We plot:
(i) \textbf{harmonic load} per edge: $\frac{1}{N}\sum_{i=1}^N \|(\mathcal{H}b^{(i)})_e\|_2$;
(ii) \textbf{diffusive activation} per vertex: $\frac{1}{N}\sum_{i=1}^N \|(z^{\star(i)})_v\|_2$;
(iii) \textbf{weight gradient magnitude}: $\|\frac{1}{N}\sum_{i=1}^N r_e^{(i)} (s_u^{(i)})^\top\|_F$.

For $\theta=0$ initialization (\figureref{fig:tension}, left column), harmonic load distributes evenly across all edges. Sheaf diffusion/inference near-uniformly spreads activations across vertices, yielding strong weight updates everywhere. For $\theta=\pi$ (\figureref{fig:tension}, right column), harmonic load concentrates only on internal edges to/from nodes $h_1$ and $h_{10}$. But diffusion is blocked at $h_1$ and $h_{10}$, starving updates on the adjacent edges because $h_2$ and $h_9$ remain inactive: contradictory feedback induces resistance to the diffusion flow.

\begin{figure}[htbp]
    \floatconts
      {fig:training}
      {\caption{\emph{Left:} validation MSE of the 10-node network under different weight initializations parametrized by $\theta$ on a single validation batch $B_{\text{val}}$ (from the same distribution as training batch $B$). Validation MSE is computed with the $y$ node unclamped (i.e., treated as a free vertex, updated during inference rather than fixed to target values). Networks with contradictory feedback ($\theta>0.4$) exhibit extremely slow convergence. \emph{Right:} Temporal evolution of harmonic load during training shows progressive ``unknotting'' dynamics for an intermediate $\theta=0.33$.}}
      {\includegraphics[width=0.98\linewidth]{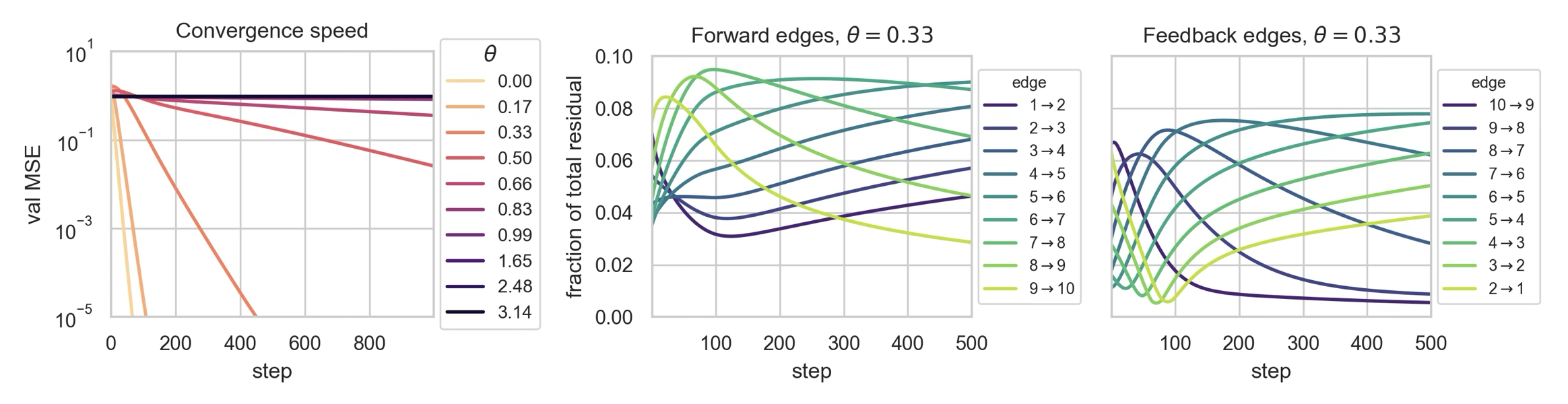}}
 \end{figure}
 \begin{figure}[tbp]
    \floatconts
      {fig:spectral}
      {\caption{Spectrum of $L=D^\top D$ vs.\ feedback angle $\theta$.
      Resonant feedback can slow inference (large $\kappa=\lambda_{\max}/\lambda_{\min}^+$) but still allows learning once converged (see \figureref{fig:training}).}}
      {\includegraphics[width=0.6\linewidth]{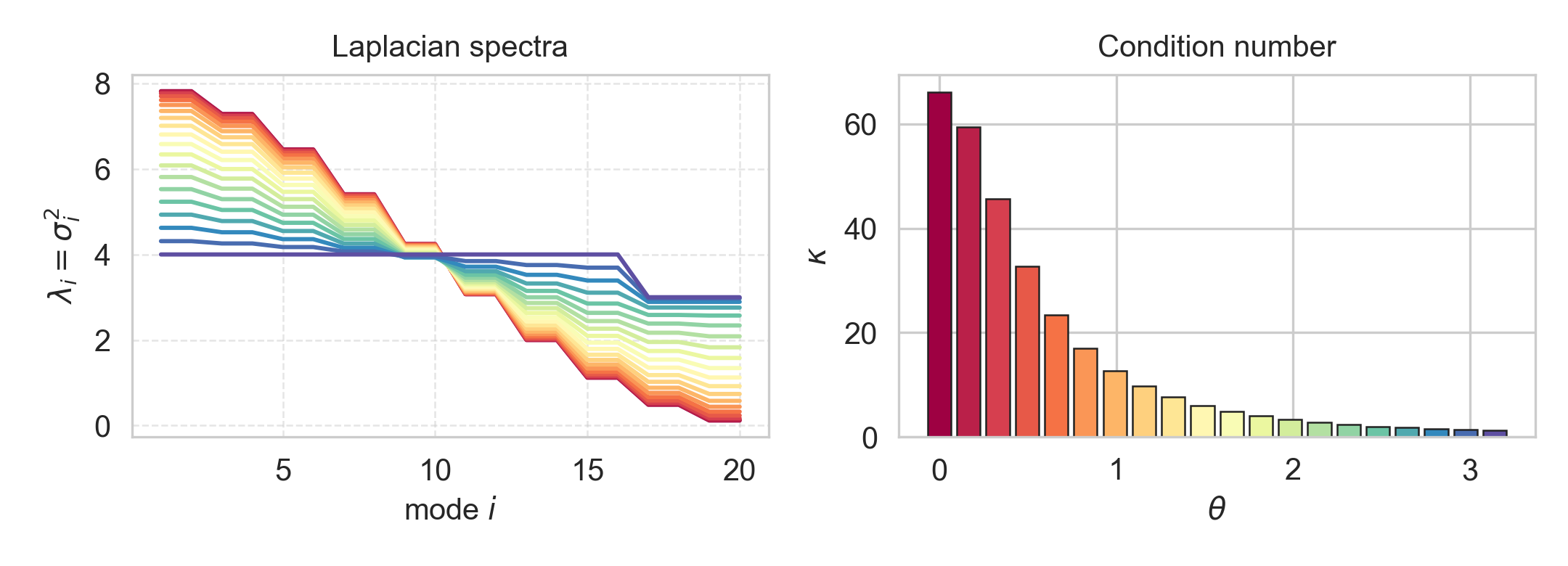}}
\end{figure}
 
To visualize training dynamics, we train for $1000$ steps at a fixed learning rate $\eta=0.1$. For each step, we solve for the optimal (and unique) $z^\star$ to study the effects of weight initialization purely on learning dynamics instead of inference dynamics. For learning dynamics, only networks with $\theta < 0.4$ achieve $\leq 0.001$ validation MSE after $1000$ steps. \figureref{fig:training} shows harmonic load evolution for an intermediate $\theta=0.33$. The choice of orthonormal initialization ensures identical per-weight norms across layers, emphasizing that learnability depends on the global wiring of the feedback loops rather than local weight scales.

Different learning rates $\eta$ can enable convergence for larger $\theta$ values, but the stark difference in convergence rates in \figureref{fig:training} demonstrates marked sensitivity to the properties of $D$ despite identical network architecture. In the appendix, we show similar analysis and results for an all-to-all predictive coding network (\appendixref{app:all-to-all}).

\section{Discussion}

Our main contribution is to show that linear predictive coding networks fit naturally within a cellular sheaf framework, recasting intuitions about prediction errors in recurrent PC networks as precise questions in the linear algebra and geometry of cellular sheaves.

\paragraph{From linear to nonlinear.} The present development can be reinterpreted by replacing each occurrence of $W$ with the Jacobian of a nonlinear layer $f$ evaluated at an equilibrium of the inference dynamics. Alternatively, introducing edgewise nonlinearities yields a nonlinear sheaf Laplacian \citep{Zaghen2024-ur, Zaghen2024-pi, Hanks2025-sp}. 

\paragraph{Inference dynamics.} Our current focus is on learnability given converged or optimal inference. In practice, and with nonlinear systems, one takes multiple gradient descent steps, where the conditioning of the inference dynamics plays a more important role. See \appendixref{app:lap-precond} for a preconditioned sheaf diffusion that preserves the minimizer set while improving conditioning.

\paragraph{Predictive coding variations.}
There exist multiple formulations of PC networks in the literature, which differ in how prediction errors are aggregated. 
Classical formulations typically associate a single prediction-error unit to each neuron or layer: all incoming predictions are combined and compared to the current state, yielding a neuron-wise squared error term in the energy \citep{Friston2005,Whittington2017,Millidge2021}. 
By contrast, our formulation uses an edge-wise factorized energy: each incident connection contributes its own squared prediction error term $\|s_v - W_e s_u\|^2$, so a neuron state is constrained simultaneously by independent consistency conditions along all of its edges. 
Architectures with multiple error channels per unit or auxiliary local losses in deep PC networks are empirically close in spirit to this edge-wise view \citep{Han2018,Tschantz2023,Oliviers2025-sc}. 
Extending our analysis to the neuron-wise setting amounts to working with a sheaf over a hypergraph, which still supports sheaf diffusion, Laplacians, and cohomology \citep{Anton2025-fg}. 
Finally, many PC models introduce precision (inverse-variance) weights on prediction errors \citep{Friston2005,Millidge2021}; in our sheaf view these correspond to modifying the inner products on edge and vertex stalks, a direction we leave open.

\section{Related Work}

The difficulty in inference and learning in predictive coding has been well-studied \citep{Qi2025-ob, Frieder2022-id}, including in the case of deep linear networks \citep{Innocenti2024-kd}. To improve learning in PC networks, recent methods have focused on depth-wise scaling \citep{Ishikawa2024-qk, Innocenti2025} for feedforward architectures. This approach has shown that appropriate scaling of weight initialization can enable stable learning in feedforward PC networks with over 100 layers. PC networks with recurrent topologies have been studied in \citet{Oliviers2025-sc, Ororbia2025-ai, Han2018}.
Standard deep linear networks have proven informative for understanding learning dynamics \citep{Saxe2013} and other phenomena \citep{Nam2025}, with observations that extend to nonlinear networks.

Cellular sheaves have been used to model opinion dynamics on social networks and other distributed systems \citep{Hansen2020-ff,Hansen2019,Hanks2025-sp}. 
Structural measures of internal tension in sheaves, such as effective resistance, have been studied in this context \citep{Hansen2018-aq}, and a connection between $1$-cochains and predictive-coding error units has been noted in \citet{Schmid2025-js}.

Sheaf neural networks generalize graph neural networks by using sheaf diffusion as the basic message-passing primitive \citep{Hansen2020-a,Bodnar2022,Barbero2022-jz,Barbero2022-kc}. 
These models operate on data sheaves over graphs (social networks, molecules, recommender graphs), and use the sheaf Laplacian to mitigate oversmoothing and to handle heterophily. 
By contrast, we apply sheaf theory to the \emph{computational graph} of a predictive coding network, with no particular input graph structure.

% \acks{[Optional acknowledgements.]}

\bibliography{references} % <-- replace with your .bib filename

@ARTICLE{Ishikawa2024-qk,
  author = {Ishikawa, Satoki and Yokota, Rio and Karakida, Ryo},
  title = {Local loss optimization in the infinite width: Stable                   parameterization of predictive coding networks and target                   propagation},
  journal = {arXiv preprint},
  year = {2024},
  note = {arXiv [cs.LG]},
  date = {2024-11-04}
}

@ARTICLE{Frieder2022-id,
  author = {Frieder, Simon and Lukasiewicz, Thomas},
  title = {(non-)convergence results for predictive coding networks},
  journal = {ICML},
  year = {2022},
  volume = {162},
  pages = {6793--6810},
  publisher = {PMLR},
  editor = {Chaudhuri, Kamalika and Jegelka, Stefanie and Song, Le and                   Szepesvari, Csaba and Niu, Gang and Sabato, Sivan},
  date = {2022}
}

@ARTICLE{Innocenti2024-kd,
  author = {Innocenti, Francesco and Achour, El Mehdi and Singh, Ryan and                   Buckley, Christopher L},
  title = {Only strict saddles in the energy landscape of predictive                   coding networks?},
  journal = {arXiv preprint},
  year = {2024},
  note = {arXiv [cs.LG]},
  date = {2024-08-21}
}

@ARTICLE{Oliviers2025-sc,
  author = {Oliviers, Gaspard and Tang, Mufeng and Bogacz, Rafal},
  title = {Bidirectional predictive coding},
  journal = {arXiv preprint},
  year = {2025},
  note = {arXiv [cs.LG]},
  date = {2025-05-29}
}

@ARTICLE{Qi2025-ob,
  author = {Qi, Chang and Forasassi, Matteo and Lukasiewicz, Thomas and                   Salvatori, Tommaso},
  title = {Towards the training of deeper predictive coding neural                   networks},
  journal = {arXiv preprint},
  year = {2025},
  note = {arXiv [cs.LG]},
  date = {2025-07-01}
}

@ARTICLE{Ororbia2025-ai,
  author = {Ororbia, Alexander and Friston, Karl and Rao, Rajesh P N},
  title = {Meta-representational predictive coding: Biomimetic self-supervised learning},
  journal = {arXiv preprint},
  year = {2025},
  note = {arXiv [cs.NE]},
  date = {2025-03-22}
}

@ARTICLE{Salvatori2025,
title = "A survey on neuro-mimetic deep learning via predictive coding",
author = "Salvatori, Tommaso and Mali, Ankur and Buckley, Christopher L and Lukasiewicz, Thomas and Rao, Rajesh P N and Friston, Karl and Ororbia, Alexander",
journal = "Neural Netw.",
publisher = "Elsevier BV",
volume =  195,
number =  108161,
pages =  108161,
year =  2025
}

@ARTICLE{Salvatori2022,
title = "Learning on arbitrary graph topologies via predictive coding",
author = "Salvatori, Tommaso and Pinchetti, Luca and Millidge, Beren and Song, Yuhang and Bao, Tianyi and Bogacz, Rafal and Lukasiewicz, Thomas",
journal = "Adv. Neural Inf. Process. Syst.",
volume =  35,
pages = "38232--38244",
year =  2022
}

@phdthesis{Curry2014-a,
  author = {Justin Curry},
  title  = {Sheaves, Cosheaves and Applications},
  school = {University of Pennsylvania},
  year   = {2014},
  type   = {PhD thesis}
}

@INPROCEEDINGS{Zaghen2024-ur,
  author = {Zaghen, Olga and Longa, Antonio and Azzolin, Steve and                 Telyatnikov, Lev and Passerini, Andrea and Liò, Pietro},
  title = {Sheaf Diffusion Goes Nonlinear: Enhancing {GNNs} with Adaptive                 Sheaf Laplacians},
  booktitle = {Geometry-grounded Representation Learning and Generative                 Modeling Workshop (GRaM) at ICML 2024},
  year = {2024},
  pages = {264--276},
  publisher = {PMLR},
  eventtitle = {Geometry-grounded Representation Learning and Generative                 Modeling Workshop (GRaM) at ICML 2024},
  date = {2024-10-12}
}

@ARTICLE{Anton2025-fg,
  author = {Anton, Ayzenberg and Thomas, Gebhart and German, Magai and                   Grigory, Solomadin},
  title = {Sheaf theory: from deep geometry to deep learning},
  journal = {arXiv preprint},
  year = {2025},
  note = {arXiv [math.AT]},
  date = {2025-02-21}
}

@ARTICLE{Zaghen2024-pi,
  author = {Zaghen, Olga},
  title = {Nonlinear sheaf diffusion in graph neural networks},
  journal = {arXiv preprint},
  year = {2024},
  note = {arXiv [cs.LG]},
  date = {2024-03-01}
}

@ARTICLE{Hansen2018-aq,
  author = {Hansen, Jakob and Ghrist, Robert},
  title = {Toward a spectral theory of cellular sheaves},
  journal = {arXiv preprint},
  year = {2018},
  note = {arXiv [math.AT]},
  date = {2018-08-04}
}

@INPROCEEDINGS{Barbero2022-kc,
  author = {Barbero, Federico and Bodnar, Cristian and de Ocáriz Borde, Haitz                Sáez and Lio, Pietro},
  title = {Sheaf Attention Networks},
  booktitle = {NeurIPS 2022 Workshop on Symmetry and Geometry in Neural                Representations},
  year = {2022},
  date = {2022-11-07}
}

@ARTICLE{Barbero2022-jz,
  author = {Barbero, Federico and Bodnar, Cristian and Borde, Haitz Sáez                   de Ocáriz and Bronstein, Michael and Veličković, Petar and                   Liò, Pietro},
  title = {Sheaf neural networks with connection Laplacians},
  journal = {arXiv preprint},
  year = {2022},
  note = {arXiv [cs.LG]},
  date = {2022-06-17}
}

@ARTICLE{Hansen2020-ff,
  author = {Hansen, Jakob and Ghrist, Robert},
  title = {Opinion dynamics on discourse sheaves},
  journal = {arXiv preprint},
  year = {2020},
  note = {arXiv [math.DS]},
  date = {2020-05-26}
}

@INPROCEEDINGS{Hansen2019,
title = "Distributed optimization with sheaf homological constraints",
author = "Hansen, Jakob and Ghrist, Robert",
booktitle = "2019 57th Annual Allerton Conference on Communication, Control, and Computing (Allerton)",
publisher = "IEEE",
year =  2019
}

@ARTICLE{Schmid2025-js,
  title        = {Applied sheaf theory for multi-agent artificial intelligence
                  (reinforcement learning) systems: A prospectus},
  author       = {Schmid, Eric},
  journal      = {arXiv preprint},
  date         = {2025-04-24},
  year         = {2025},
  eprinttype   = {arXiv},
  eprintclass  = {math.OC}
}

@ARTICLE{Hanks2025-sp,
  author = {Hanks, Tyler and Riess, Hans and Cohen, Samuel and Gross,                   Trevor and Hale, Matthew and Fairbanks, James},
  title = {Distributed multi-agent coordination over cellular sheaves},
  journal = {arXiv preprint},
  year = {2025},
  note = {arXiv [math.OC]},
  date = {2025-04-02}
}

@ARTICLE{Innocenti2025,
title = "$\mu${PC}: Scaling predictive coding to {100+} layer networks",
author = "Innocenti, Francesco and Achour, El Mehdi and Buckley, Christopher L",
journal = "arXiv [cs.LG]",
year =  2025,
archivePrefix = "arXiv",
primaryClass = "cs.LG"
}

@ARTICLE{Saxe2013,
title = "Exact solutions to the nonlinear dynamics of learning in deep linear neural networks",
author = "Saxe, Andrew M and McClelland, James L and Ganguli, Surya",
journal = "arXiv [cs.NE]",
year =  2013,
archivePrefix = "arXiv",
primaryClass = "cs.NE"
}

@ARTICLE{Nam2025,
title = "Position: Solve layerwise linear models first to understand neural dynamical phenomena (neural collapse, emergence, lazy/rich regime, and grokking)",
author = "Nam, Yoonsoo and Lee, Seok Hyeong and Domine, Clementine and Park, Yea Chan and London, Charles and Choi, Wonyl and Goring, Niclas and Lee, Seungjai",
journal = "arXiv [stat.ML]",
year =  2025,
archivePrefix = "arXiv",
primaryClass = "stat.ML"
}

@ARTICLE{Salvatori2022-a,
title = "A stable, fast, and fully automatic learning algorithm for predictive coding networks",
author = "Salvatori, Tommaso and Song, Yuhang and Yordanov, Yordan and Millidge, Beren and Xu, Zhenghua and Sha, Lei and Emde, Cornelius and Bogacz, Rafal and Lukasiewicz, Thomas",
journal = "arXiv [cs.NE]",
year =  2022,
archivePrefix = "arXiv",
primaryClass = "cs.NE"
}

@BOOK{Rosiak2022-a,
title = "Sheaf Theory through examples",
author = "Rosiak, Daniel",
publisher = "The MIT Press",
year =  2022
}

@ARTICLE{Whittington2017,
title = "An approximation of the error backpropagation algorithm in a predictive coding network with local Hebbian synaptic plasticity",
author = "Whittington, James C R and Bogacz, Rafal",
journal = "Neural Comput.",
publisher = "MIT Press - Journals",
volume =  29,
number =  5,
pages = "1229--1262",
year =  2017
}

@ARTICLE{Millidge2021,
title = "Predictive coding: A theoretical and experimental review",
author = "Millidge, Beren and Seth, Anil and Buckley, Christopher L",
journal = "arXiv [cs.AI]",
year =  2021,
archivePrefix = "arXiv",
primaryClass = "cs.AI"
}

@ARTICLE{Friston2005,
title = "A theory of cortical responses",
author = "Friston, Karl",
journal = "Philos. Trans. R. Soc. Lond. B Biol. Sci.",
publisher = "The Royal Society",
volume =  360,
number =  1456,
pages = "815--836",
year =  2005
}

@ARTICLE{Han2018,
title = "Deep Predictive Coding Network with Local Recurrent Processing for Object Recognition",
author = "Han, Kuan and Wen, Haiguang and Zhang, Yizhen and Fu, Di and Culurciello, Eugenio and Liu, Zhongming",
journal = "Advances in Neural Information Processing Systems",
volume =  31,
year =  2018
}

@ARTICLE{Tschantz2023,
title = "Hybrid predictive coding: Inferring, fast and slow",
author = "Tschantz, Alexander and Millidge, Beren and Seth, Anil K and Buckley, Christopher L",
journal = "PLoS Comput. Biol.",
volume =  19,
number =  8,
pages = "e1011280",
year =  2023
}

@ARTICLE{Hansen2020-a,
title = "Sheaf Neural Networks",
author = "Hansen, Jakob and Gebhart, Thomas",
journal = "arXiv [cs.LG]",
year =  2020,
archivePrefix = "arXiv",
primaryClass = "cs.LG"
}

@INPROCEEDINGS{Bodnar2022,
title = "Neural Sheaf Diffusion: A Topological Perspective on Heterophily and Oversmoothing in {GNNs}",
author = "Bodnar, Cristian and Di Giovanni, Francesco and Chamberlain, Benjamin Paul and Lio, Pietro and Bronstein, Michael M",
booktitle = "Advances in Neural Information Processing Systems",
year =  2022
}

\appendix

\section{Laplacian preconditioner}

\label{app:lap-precond}
Following \citet{Hansen2018-aq}, for the relative least-squares problem $E_{\mathrm{rel}}(z)=\frac{1}{2}\|Dz+b\|^2$ with $L_{\mathrm{rel}}=D^\top D$, use the block-Jacobi preconditioner
\begin{equation}
M := \mathrm{blkdiag}(L_{\mathrm{rel}}) \quad (\text{vertex blocks}).
\end{equation}
Run preconditioned gradient descent
\begin{equation}
z^{(t+1)} = z^{(t)}-\eta M^{-1}D^\top\big(Dz^{(t)}+b\big).
\end{equation}
Fixed points satisfy $M^{-1}D^\top(Dz+b)=0 \iff D^\top(Dz+b)=0$, so the minimizer set is unchanged. If $L_{\mathrm{rel}}$ is nonsingular, the unique optimizer remains $z^\star=-L_{\mathrm{rel}}^{-1}D^\top b$. If $\ker D\neq\{0\}$, the affine solution set is the same and the iteration selects the $M$-minimum-norm representative within it.

Equivalently, endow the vertices with the inner product $\langle u,\,v\rangle_M=u^\top Mv$ and keep the standard inner product on edges. Then the adjoint of the coboundary is $D^*_M=M^{-1}D^\top$, so the preconditioned iteration is steepest descent for the same objective in the $M$-metric. The diffusion operator becomes $M^{-1}L_{\mathrm{rel}}$, and under the change of variables $y=M^{1/2}z$ this takes the normalized form $M^{-1/2}L_{\mathrm{rel}}M^{-1/2}$.

\section{Gauss--Newton per-edge learning rates}
\label{sec:edge-lrs-canonical}

At the inference optimum, the residual on edge $e=(u\!\to\!v)$ and source activation at $u$ depend linearly on the boundary $b$:
\begin{align}
r_e^\star &= (R_e \mathcal{H})\,b, &
s_u^\star &= - (S_u \mathcal{G})\,b,
\end{align}
where $R_e,S_u$ extract edge and vertex blocks respectively.

For $b\sim\mathcal N(0,\Sigma_b)$ on $C^1$, the Gauss--Newton method requires the source covariance
\begin{align}
\Sigma_{s_u} = \mathbb{E}\!\left[s_u^\star (s_u^\star)^\top\right]
= S_u \mathcal{G}\,\Sigma_b\, \mathcal{G}^\top S_u^\top.
\label{eq:sigma_su}
\end{align}
When $\Sigma_b=\sigma^2 I$,
\begin{align}
\Sigma_{s_u} = \sigma^2\, S_u L_{\mathrm{rel}}^\dagger S_u^\top,
\label{eq:blockL}
\end{align}
since $\mathcal{G} \mathcal{G}^\top = L_{\mathrm{rel}}^\dagger$.

\paragraph{Per-edge update rule.}
For mini-batch gradient $G_e := \sum_{i=1}^N r_e^{(i)} (s_u^{(i)})^\top$ at edge $e=(u\!\to\!v)$, the update is
\begin{equation}
W_e \leftarrow W_e - \gamma G_e \big(\Sigma_{s_u}+\varepsilon I\big)^{-1},
\quad 0<\gamma<2,\ \varepsilon>0.
\label{eq:gn}
\end{equation}
This uses only source statistics, omitting residual-side normalization.

\paragraph{Scalar fallback.}
If a matrix preconditioner is impractical, use a per-source scalar rate:
\begin{equation}
\eta_e^{\mathrm{spec}} \;=\; \frac{\gamma}{\varepsilon + \lambda_{\max}(\Sigma_{s_u})},
\qquad
W_e \leftarrow W_e - \eta_e^{\mathrm{spec}} G_e.
\label{eq:scalar}
\end{equation}

\paragraph{Estimating $\Sigma_{s_u}$ without explicit $\mathcal{G},\mathcal{H}$.}
Hutchinson probes: draw $\xi_k\sim\mathcal N(0,\Sigma_b)$ in $C^1$ and solve
\begin{align}
L_{\mathrm{rel}} y_k \;=\; D^\top \xi_k \quad \text{(or $(L_{\mathrm{rel}}+\lambda I)y_k=D^\top \xi_k$ if $L_{\mathrm{rel}}$ is singular)},
\end{align}
then accumulate
\begin{align}
\widehat\Sigma_{s_u} \;\approx\; \frac{1}{q}\sum_{k=1}^q (S_u y_k)(S_u y_k)^\top.
\label{eq:hutch}
\end{align}

\paragraph{Spectral analysis.}
The selection operator $S_u$ groups all edges from source vertex $u$ under the same covariance $\Sigma_{s_u}$. With eigendecomposition $L_{\mathrm{rel}}=U\Lambda U^\top$,
\begin{align}
\Sigma_{s_u} = S_u U \Lambda_{\mathrm{rel}}^\dagger U^\top S_u^\top
\quad(\text{whitened case}),
\end{align}
where small eigenvalues of $L_{\mathrm{rel}}$ inflate $\Sigma_{s_u}$ and reduce the effective step size through $(\Sigma_{s_u}+\varepsilon I)^{-1}$. Poorly conditioned modes thus enforce conservative updates. The operator $\mathcal{G}$ determines vertex reachability under diffusion: vertices with small $s_u^\star$ yield weak gradients. The harmonic projector $\mathcal{H}$ filters the gradient through $r_e^\star=(R_e \mathcal{H})b$; when $\mathcal{H}$ concentrates residuals on edges from weakly activated sources, learning can stall despite preconditioning.

\begin{figure}[tbp]
    \floatconts
      {fig:rnn}
      {\caption{All-to-all variant: harmonic load concentrates on edges incident to the penultimate node $h_f$, while activations are weaker elsewhere, diminishing updates on $h_i\!\to\!h_f$ edges.}
      }
      {\includegraphics[width=0.98\linewidth]{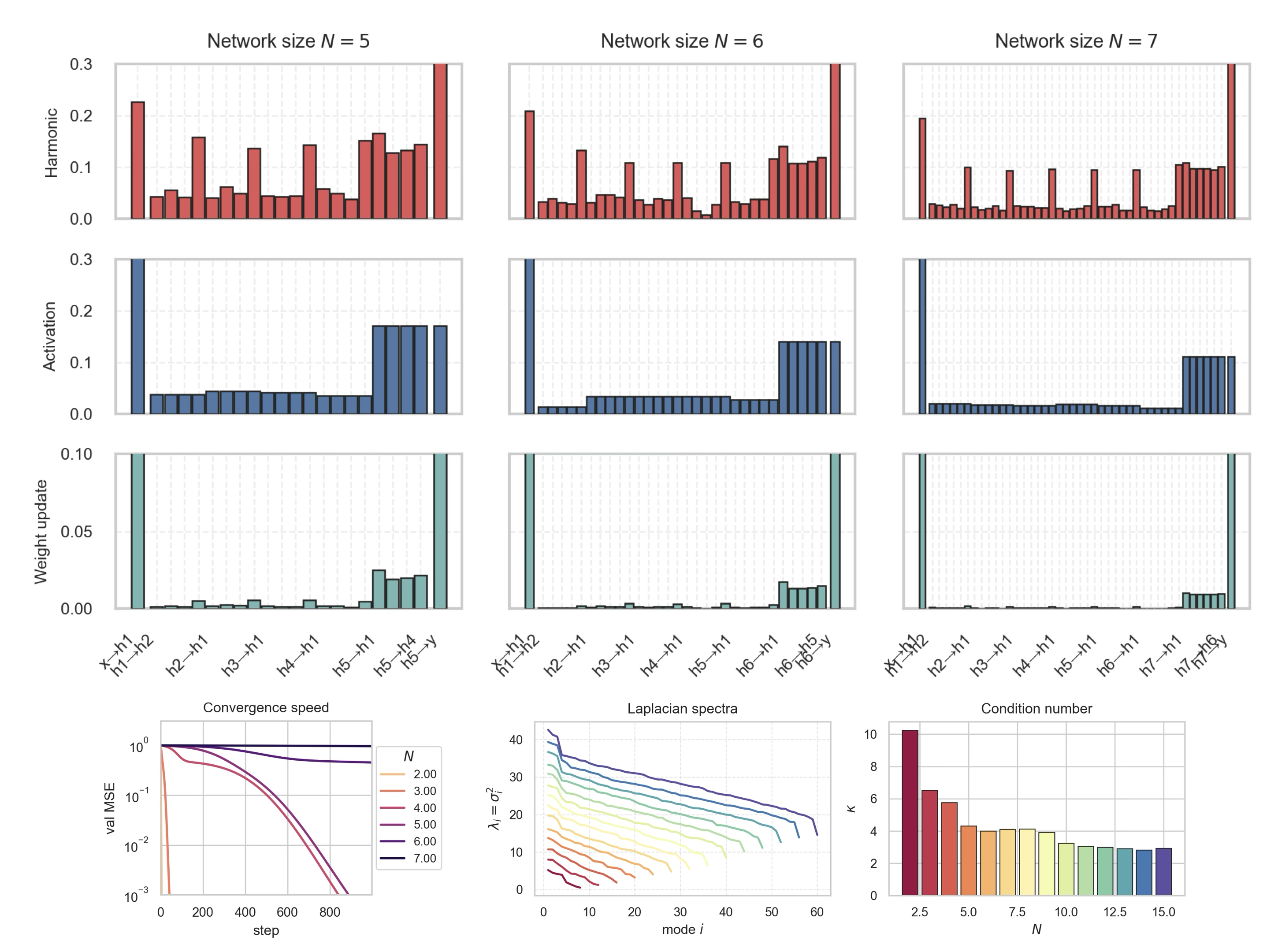}}
\end{figure}

\section{All-to-all network}
\label{app:all-to-all}
We observe similar behavior in an all-to-all network. In \figureref{fig:rnn}, we show training dynamics of an all-to-all network, where one $x$ and $y$ nodes connect to one of the $h_i$ nodes each, and every $h_i$ node connects to every other node (bidirectionally) with orthonormally initialized weights, with each $h_i$ 4-dimensional and $x$ and $y$ 2-dimensional. Harmonic load concentrates primarily on edges connected to the penultimate node (that which is connected to $y$, call it $h_f$), while activations are weaker on non-penultimate nodes, resulting in weaker updates on $h_i\rightarrow h_f$ edges. For network sizes ranging from $2$ to $15$ only networks of size $5$ or less achieve $\leq 0.001$ validation MSE after $1000$ steps, again highlighting the effect of ``internal tension'' on convergence rates.

\end{document}